Ziny Flikop

# Bounded-Input Bounded-Predefined-Control Bounded-Output

7 Fisher Ln. Levittown, NY , 11756

Abstract
The paper is an attempt to generalize a methodology [1; 93] and [2, 3; 95], which is similar to the bounded-input bounded-output (BIBO) method currently widely used [4-9] for the system stability studies. The presented in [1, 2] methodology allows decomposition of input space into bounded subspaces and defining for each subspace its bounding surface. It also defines a corresponding predefined control, which maps any point of a bounded input into a desired bounded output subspace. This methodology was improved [3] by providing a mechanism for the fast defining a bounded surface. This paper combines BIBO with bounded-control [10-14] and presents enhanced "bounded-input bounded-predefined-control bounded-output" (BIBPCBO) approach, which provides adaptability feature to the control and allows transferring of a controlled system along a suboptimal trajectory.

## Introduction

Today predefined (predefined) control becomes a very powerful methodology used for diverse applications [15-20]. At the same time, "bounded-input bounded-output" (BIBO) approach is widely used for the stability analysis of different complex systems [4-9]. In [1.2 and 3] we have presented an approach, which combines BIBO with the predefined control. In this paper, we add to it a bounded-control [4-9] aspect and new enhanced abilities. Considering that papers [1, 2 and 3] were published in difficult to find today publications and to maintain integrity of analysis below, we repeat in a very compressed way main ideas presented in those papers. To provide a maximum simplicity in our analysis we avoid any mathematical inclusions in the text. (Reader, who wants to see a more strict analysis of the proposed methodology and evaluate numerical results of simulation a complex non-liner system, can see [1, 2 and 3].)

 Now, let us consider a controllable system (object, process, etc.) as a system that is determined by a set of input parameters, a set of output parameters, and some mechanism that allows mapping of valuable inputs into acceptable outputs. A system can be either linear or non-linear.  In the case when some system parameters are fuzzy, for example, color, taste, etc., they can be defined via digital directional sequences.

We can define a system state as a combination of some system parameters, i.e., as a point into a multidimensional conditional input space. Since in the future we will use a solution (predefined control), instead of the term "solution", we will also use the term "predefined control" or just "control". Some controls are relatively simple. Others are more complicated. They can include predefined controls ( solutions)
as initiators and closed-loop controls as executors [16, 19]. Some controls are defined in the multidimensional control space. Predefined controls can consist of  multi-steps with closed-loop inclusions.

For the design of a predefined control system based on BIBO, system engineer can use either real object (process) or its mathematical model. For the simplicity, we will assume that for our following discussion we have a mathematical model of an object.

Let us start with preliminary study of a problem. In the very beginning, for a particular object state (we name it an "origin" point in the input space) we can found via optimization a "best" possible control that maps the "origin" point into the "best" desired output (in the one-dimensional case see Fig. 1). If for every possible input we will define the "best" control that provides us with the "best" output, then we finish with an infinite number of controls (in the one-dimensional case, see Fig. 2).

However, as system designers, we can define a bounded-output space inside of which system output can fluctuate and, in spite of these fluctuations, it can still be considered as acceptable. Now we can have two extreme situations. Namely, for one input we will have a set of "acceptable" controls (in the one-dimensional case, see Fig 3), or for one "best" control we will have an acceptable input area (in the one-dimensional case, see Fig 4).

It is possible to have an intermediate solution when instead of one "best" control; we can find a set of controls that satisfies desired output conditions. However, the price we pay for this is a shrinking acceptable input area (in the one-dimensional case, see Fig 5).

When we use a fixed "best" control, the more input deviates from the "origin" point, the more output departs from its "best" value. It means that with too-wide deviations of input a "goodness" of a found "best" control can deteriorate to the extent that departed input will be mapped outside of a bounded-output space, i.e., control becomes either useless or, in the worst case, counterproductive. Border points in the input space that the "best" control cannot map into a desired output are the cutoff points for this control (in the one-dimensional case, see Fig 4). In the multidimensional case, cutoff points create an outer surface of an area in the input space. In other words, these points create a bounded input subspace. Any point in this subspace will be mapped by a "best" control into a desired bounded output.

Implementation of this approach requires two phases: learning and execution. Let us start with an analysis of the learning phase in more detail. We will do our analysis of a generic system described by an input, output, and control. All or part of these parameters can be multidimensional and non-liner. It is obvious that dimensionality of input space, control space, and output space can be different.

Learning

1. As a first step, for the given input represented by the "origin" point in the multidimensional input space, we will find via optimization a "best" control that maps input into the "best" desired output. This multidimensional optimization process is complicated; however, result of it is similar to the case shown in Fig. 1. Then, for a bounded-output space, we can identify a bounded input subspace via reverse mapping. The "best" control will map any point of this subspace into bounded output. We will do reverse mapping in few steps:

2. In the very beginning, we normalize an input space at the "origin" point by defining a measure in some units for each input parameter (dimension). For example, a unit of temperature can be 1/100 of a maximal considered temperature; the taste of a fruit can be 1/100 of maximal taste units, and so on.

3. This normalization allows us define a directional ray [3] that originates at the "origin" point and is going in the input space in a random direction. We can achieve this by multiplying each dimensional unit by corresponding to it random number.
4. Now, we can start to move the input point away from the "origin" along a randomly selected ray. As a result, the quality of the output will steadily diminish until we arrive at an output border. The input point that the "best" control maps into this border is a cutoff input point. Considering that a ray is one-dimensional, we can use a very fast one-dimensional optimization algorithm to find a cutoff point for the combination: direction, "best" control and an output border. Coordinates of a found cutoff point are memorized.
5. By using another set of random numbers, we can define a new directional ray that originates in the same "origin" point, repeat the optimization procedure and memorize a new cutoff point.
6. After we repeat step 5 a number of times, we will have a set of cutoff points that determine a boundary of the input subspace, any point inside of which can be mapped by the "best" control inside of a bounded output space.
7. The cutoff points allow us to define preliminarily by some polynomial a multidimensional surface that bounds the found input subspace. In this, we can rely on a hypothesis that, with the exception of systems designed by man, this surface is convex.

In addition, the bounded input space restrained by this surface does not contain subspaces input points in which cannot be mapped by analyzed control into a bounded output. (More detailed analysis of this problem presented in [2].) We rely on these hypotheses until otherwise is proven. Polynomial description of the surface can be done by using a "Fit" function of [21], or another fitting algorithm. The result of a fitting provides us with information about the polynomial and the value of the least square error corresponding to it. Using this polynomial allows us to speed up our process, since now we can continue selection of testing input points that are very close to the bounding surface.

These points can be randomly selected one-by-one and each point is considered as the "origin" for the procedure described in steps 3 and 4. Each additional cutoff point is used to correct the polynomial and recalculate the value of a new least square error. We continue this procedure until this error is stabilized. Then we terminate the procedure and consider that the polynomial is found. For the two-dimensional input, Fig. 6 illustrates this procedure. After the polynomial is found, we memorize a trio: a polynomial that defines the bounded input subspace, the "best" control that maps any input that belongs to this subspace into a bounded output, and the bounded output itself.

The approach described in steps 1-7 has been successfully verified on the model of a telecommunications network [1, 2, and 3]. This non-liner model represents a system with 31-dimensional input, 43-dimensional control and one-dimensional output. For the polynomial fitting was used a "Fit" function [21].

8. Now, we can try to add some flexibility to the already found "best" control. To do so, we, for the same "best" output and bounded output space, will choose an input point located inside of already defined bounded input subspace. We consider it as a new "origin" point. If we repeat steps 3 through 7, we can find another "best" control and another bounded input subspace that is intersecting with the previously found input subspace. Now, either of the controls will map all input points that belong to the intersection into a bounded output. We can repeat step 8 a number of times and define such control subspace any control in which maps any point of an intersected input subspace into a desired bounded output space.

9. We also can try to repeat steps 1-7 for the same "best" output and bounded output space, however now we will choose inputs that are outside of intersected input subspace however, located in a close proximity to it. If input subspaces for these points are found to be intersecting with the previously found input subspace intersection, we can accept newly found controls as valid controls for the bounded output space. This will allow us expand control space found in 8.

10. We can repeat step 8-9 as many times as our time resources allow. As a result, we will have a not empty intersected input subspace all points of which can be mapped into bounded output by any control of a found control subspace. We hypothesize that this control space is convex, and determine its bounding surface by a polynomial. Now we will define a bounded input subspace as an intersection of input subspaces and a set of controls via a bounded control subspace.

Let us modify our memorized trio. Since any control of a bounded control subspace maps any input of a bounded input subspace into the bounded output space, we memorize "bounded-input-subspace bounded-control-subspace bounded-output-space" (BIBCBO) solution.

Steps 1-10 can be repeated for completely different points in the input space. As a result, we decompose an input space on controllable subspaces and will have a library of solutions for different operational situations. A set of memorized trios allows us to execute a system control in real time.

In addition, if we maintain the same "origin" point, but move the "best" output and bounded output along some path, then the result of multiple repetitions of steps 1, 3-10 will allow us transfer an analyzed system from one state to the other state along a desired suboptimal trajectory.

If we maintain the same "best" output and desired bounded output, but move the "origin" point in some direction, then the result of steps 1, 3-10 will let us equip analyzed systems with adaptive abilities.

Educational time can be significantly reduced by utilization of an available expertise. We can mimic an educational processes that humans and some animals use to transfer their own expertise to children. In complex cases, experts can teach, for example, a robot to do different tasks. The experience of a test pilot can be used for the development recovery procedures in case of emergencies.

Possible applications of the proposed methodology

1. It becomes feasible to create fail-proof systems. To do so, we can limit system control by permitted bounded control subspace and, as a result, prevent some accidents. This approach, for example, allows prevention of airplane crashes, similar to the crash of the Egyptian airliner, or the Chernobyl accident.

2. We can prepare in advance a set of flexible solutions for a robot. A robot will execute these solutions depending on the required task and environmental conditions. In addition, a robot can adjust its reactions upon changing environmental conditions or output requirements. This is a very important ability in cases when a robot is designed to work independently without human intervention.

3. It becomes possible to develop in advance a set of flexible recovery procedures for the natural disasters.
4. Systems (objects), for example aviation engines, can be transferred from one state to another via a sub-optimal trajectory.
5. The experience of an expert can be reused for the development of recovery procedures in case of emergencies.
With some imagination, one can significantly expand this list of possible applications.

Execution (Predefined Control)

Execution of the proposed approach is relatively straightforward. Control system monitors input conditions of the controlled system and defines to which bounded input subspace the current input belongs. If input belongs to one of the subspaces defined during the learning process, then control corresponding to this subspace is executed.

In the cases of recognition, information about object (sound) parameters is used to define an object via a library of solutions. A found solution gives us a probability that recognition is correct.

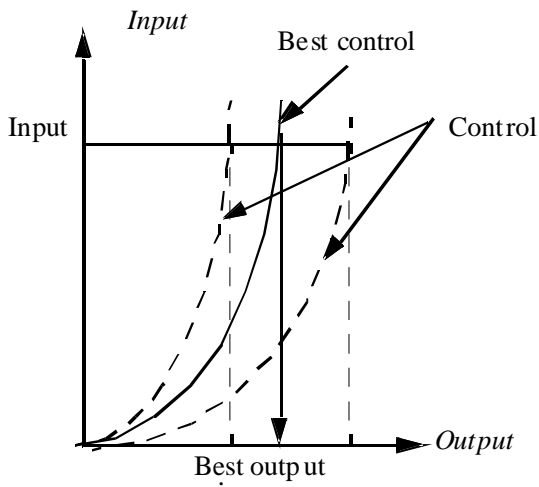

Fig 1.

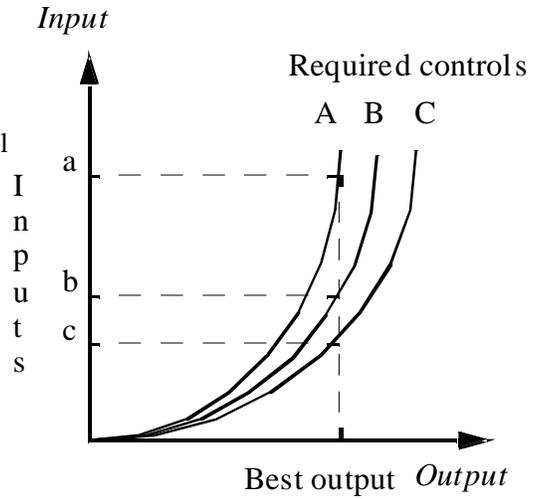

Fig 2.

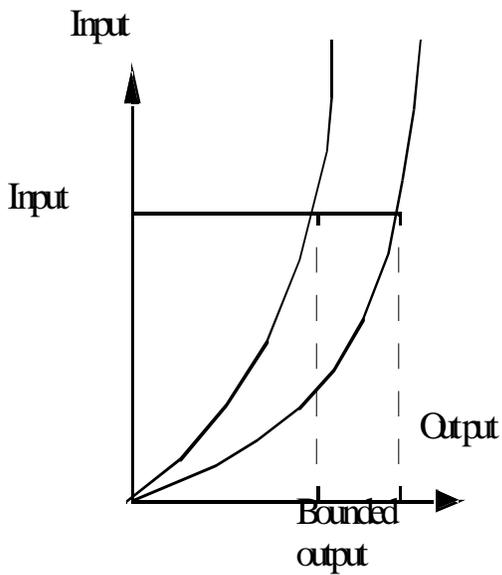

Fig 3

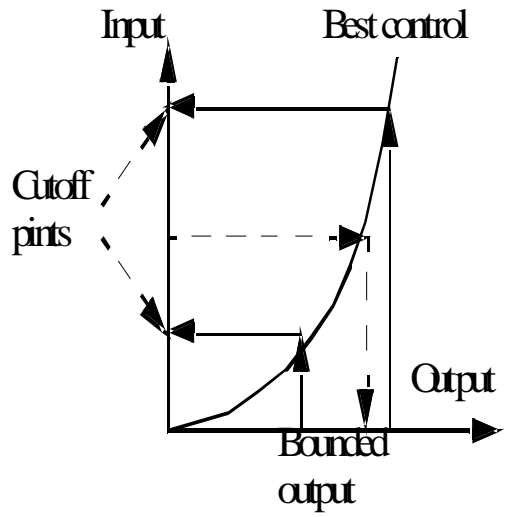

Fig 4

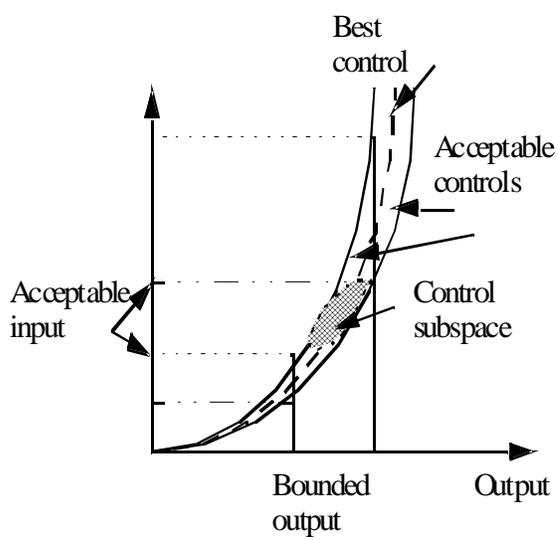
Fig. 5

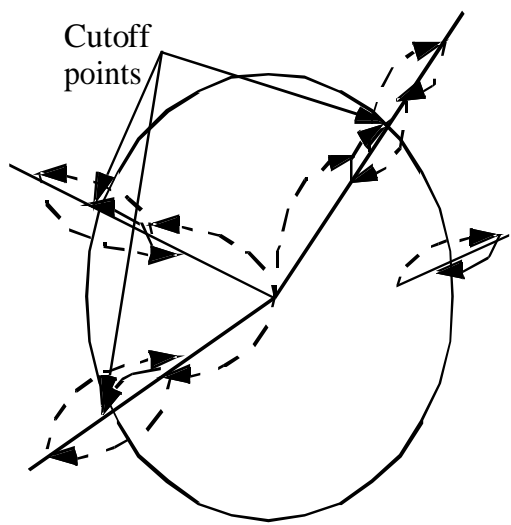
Fig 6.